\documentclass[conference]{IEEEtran}
\usepackage{cite}
\usepackage{amsmath,amssymb,amsfonts}
\usepackage{algorithmic}
\usepackage{graphicx}
\usepackage{textcomp}
\usepackage{xcolor}
\def\BibTeX{{\rm B\kern-.05em{\sc i\kern-.025em b}\kern-.08em
    T\kern-.1667em\lower.7ex\hbox{E}\kern-.125emX}}
\begin{document}

\title{Heuristic Hyperparameter Choice for Image Anomaly Detection\\
\thanks{Identify applicable funding agency here. If none, delete this.}
}

\author{\IEEEauthorblockN{Zeyu Jiang}
\IEEEauthorblockA{\textit{{Centre for mathematical morphology}} \\
\textit{Mines Paris - PSL}\\
Paris, France \\
zeyu-valentin.jiang@etu.minesparis.psl.eu}
\and
\IEEEauthorblockN{João P. C. Bertoldo}
\IEEEauthorblockA{{\textit{Centre for mathematical morphology}} \\
\textit{Mines Paris - PSL}\\
Paris, France \\
jpcbertoldo@minesparis.psl.eu}
\and
\IEEEauthorblockN{Etienne Decencière}
\IEEEauthorblockA{\textit{{Centre for mathematical morphology}} \\
\textit{Mines Paris - PSL}\\
Paris, France \\
etienne.decenciere@mines-paristech.fr}
}

\maketitle

\begin{abstract}
Anomaly detection (AD) in images is a fundamental
computer vision problem by deep learning neural network to identify images deviating significantly from normality.
The deep features extracted from pretrained models have been
proved to be essential for AD based on multivariate Gaussian distribution analysis. 
However, since models are usually pretrained on a large dataset for classification tasks such as ImageNet, they might produce lots of redundant features for AD, which increases computational cost and degrades the performance.
We aim to do the dimension reduction of Negated
Principal Component Analysis (NPCA) for these features. So we proposed some heuristic to choose hyperparameter of NPCA algorithm for getting as fewer components of features as possible while ensuring a good performance.
\end{abstract}


\section{Introduction}
Anomaly detection task (AD task) is generally understood to be the identification of rare items, events or observations which deviate significantly from the majority of the data and do not conform to a well defined notion of normal behaviour.

Anomaly detection is applicable in a very large number and variety of domains, and is an important subarea of machine learning. For example, for the face presentation attack detection in biometric
systems, attack accesses can be considered abnormal and should be distinguished from the real-access data. In the medical field, AD might refer to identifying a disease or locating the lesion region on the medical images. Specifically, in industrial fields, AD
is related to detect abnormal real-time power consumption or defects in industrial objects so it can help to reduce costs and improve efficiency.

A large group of AD methods based on binary classification can be classified as the supervised learning strategy requiring
both normal and abnormal samples during training. However,
supervised learning methods are not realistic in many practical applications because of the scarcity of labeled training
data and the class imbalance (i.e., the abnormal samples are
often more difficult to collect than normal ones). Also, other semi-supervised or unsupervised learning methods where a neural network is fitted from scratch might encounter the overfitting issue because of insufficient training data.

Previous works have proposed transfer learning to circumvent these limitations, which involves pretraining a classifier on
a large dataset (e.g., ImageNet) and fine-tuning the classifier
using a smaller labeled AD dataset related to specific tasks.

In our paper, AD is applied to detect the defaults in objects and textures (particularly MVTec AD dataset). An abnormal observation is defined as an image for which the items has some defaults such as being broken or being painted. Our strategy is to study the multivariate gaussian (MVG) distribution of pretrained deep learning features. Based on this, we proposed a heuristic to choose as few components of features as possible but still have a good performance of AD.

\section{Related work}
In recent years, many articles have been published in the field of anomaly detection applied on the MVTec AD dataset, a comprehensive real-world
dataset\cite{b1}. There are 15 different classes of high-resolution color images of industrial objects or textures, totally 5354 images. Fig.1\cite{b1} shows some of them. It contains normal, i.e., defect-free images intended for training and images with anomalies intended for testing. The anomalies manifest themselves in the form of over 70 different types of defects such as
scratches, dents, contaminations, and various structural changes.

\begin{figure}[!t]
\centering
\includegraphics[scale=0.25]{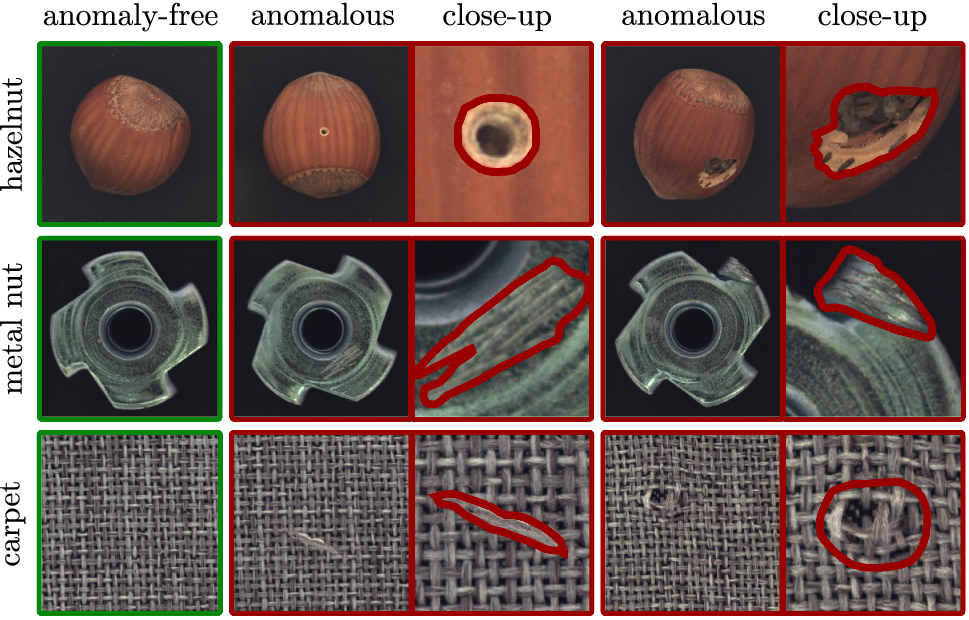}
\caption{Two objects (hazelnut and metal nut) and one texture (carpet)
from the MVTec Anomaly Detection dataset. For each of them, one
defect-free image and two images that contain anomalies are displayed.
Anomalous regions are highlighted in close-up figures together with
their pixel-precise ground truth labels. The dataset contains objects and
textures from several domains and covers various anomalies that differ
in attributes such as size, color, and structure}
\label{fig:1}
\end{figure}

Reference \cite{b2} models the distribution of normal data in deep feature representations learned from ImageNet via a MVG and their approach achieves a new state of the art in AD on MVTec AD dataset.

\begin{figure}[!t]
\centering
\includegraphics[scale=0.4]{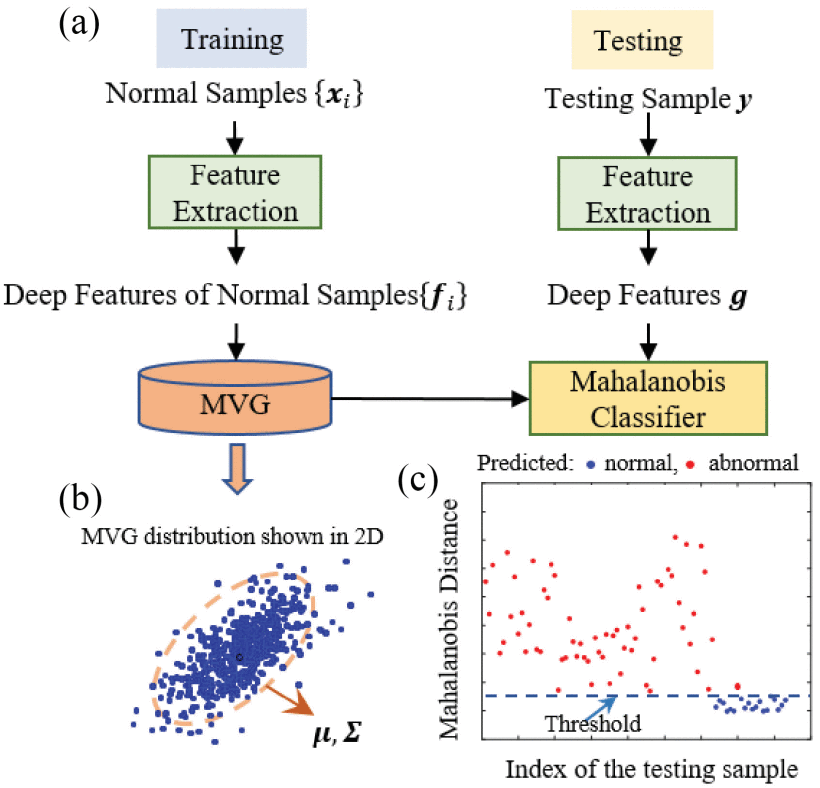}
\caption{AD based on Gaussian discriminative analysis and deep features.
(a) Framework of the method. (b) Schematic of an MVG distribution. (c) Classifier based on Mahalanobis distances}
\label{fig:1}
\end{figure}

Reference \cite{b3} has also applied this approach in Fig.2 \cite{b3} and proposed two methods to select the useful deep features for the AD task instead of applying all the features extracted. The first is to identify and select the most effective stage of layer and the second is to do the dimension reduction of features with subspace decomposition.
For the second method, they proposed ratio of eigenvalue heuristic for the choice of heuristic parameter $\tilde{k}$ to do the dimension reduction,
\begin{equation}
\tilde{k} = \arg \max_{1 \le m \le d-1} \frac{\lambda_{m}}{\lambda_{m+1}}
\label{eq}
\end{equation}
where $d$ is the dimension of covariance matrix and $(\lambda_1, \lambda_2, \lambda_3, \cdots, \lambda_d$) is the list of eigenvalues of covariance matrix with $ \lambda_1 < \lambda_2 < \lambda_3 < \cdots < \lambda_d $. And they showed the mean of performance for all categories.

We extend the heuristic presented in \cite{b3} by comparing it with the performance with all possible parameters of dimension reduction on each stage of layer, identifying its effectiveness. And then we propose an alternative heuristic to select the deep feature by determining the parameter of dimension reduction algorithm.

\section{Method}

Inspired by Lin's work, we have applied the transfer learning with the model that is of architecture EfficientNet-B4 trained on ImageNet dataset. The general idea of heuristic choice of parameter $k$ of Negated Principal Component Analysis (NPCA) is shown below.

1. Train the pretrained model on MVTec AD dataset.

2. Extract features from all the stages of layers in the model and named them respectively from feature.0 to feature.8.

3. Fit the features extracted to a Multivariate Gaussian (MVG) distribution by determining two parameters, the mean of features $\mu$ and covariance matrix of features $\Sigma$.

4. Apply the whitening operation on the features.

5. Apply dimension reduction of NPCA with parameter $k$ on each features and plot the receiver operating characteristic (roc) curve.
Here, to detect the abnormal images, we calculate the Euclidean distance $D$ of feature and set the threshold $D_t$. During testing the images inputted, we consider it as normal images if they have the distance $D<D_t$ while we consider it abnormal if $D>D_t$.

6. Plot the area under 
roc (auroc) with respect to all possible $k$ on each features and detemine the optimal parameter $k^*$.

7. Propose the heuristic to choose the heuristic parameter $\tilde{k}$ of NPCA and compare it to the optimal one $k^*$.

Next, we give an overview with Multivariate Gaussian distribution (MVG), Negated Principal Component Analysis (NPCA) and whitening transformation.
\subsection{Multivariate Gaussian distribution}
The Probability Density Function (PDF) of MVG for the feature extracted  $\mathbf{x}$ of image is given as
\begin{equation}
\varphi_{\mu, \Sigma}(\mathbf{x}):=\frac{1}{\sqrt{(2 \pi)^{d}|\operatorname{det} \Sigma|}} e^{-\frac{1}{2}(\mathbf{x}-\mu)^{\top} \Sigma^{-1}(\mathbf{x}-\mu)}
\label{eq}
\end{equation}
with $d$ being the number of dimension of the features, $\boldsymbol{\mu} \in \mathbb{R}^{d}$ being the mean vector, and $\Sigma \in \mathbb{R}^{d \times d}$ being the symmetric covariance matrix of the distribution.

We prefer to use the Mahalanobis distance as the score instead of common euclidean distance because we want to eliminate effects of covariance between the components of feature. for the given test data $\mathbf{x}$, it is defined as,
\begin{equation}
M(\mathbf{x})=\sqrt{(\mathbf{x}-\mu)^{\top} \Sigma^{-1}(\mathbf{x}-\mu)} .
\label{eq}
\end{equation}

\subsection{Negated Principal Component Analysis}
Principal Component Analysis (PCA) is a widely used algorithm for data dimension reduction. The main idea of PCA is to map n-dimensional features to k-dimensions. This new k-dimensional feature is called principal component of the original n-dimensional feature.

\begin{figure}[!h]
\centering
\includegraphics[scale=0.3]{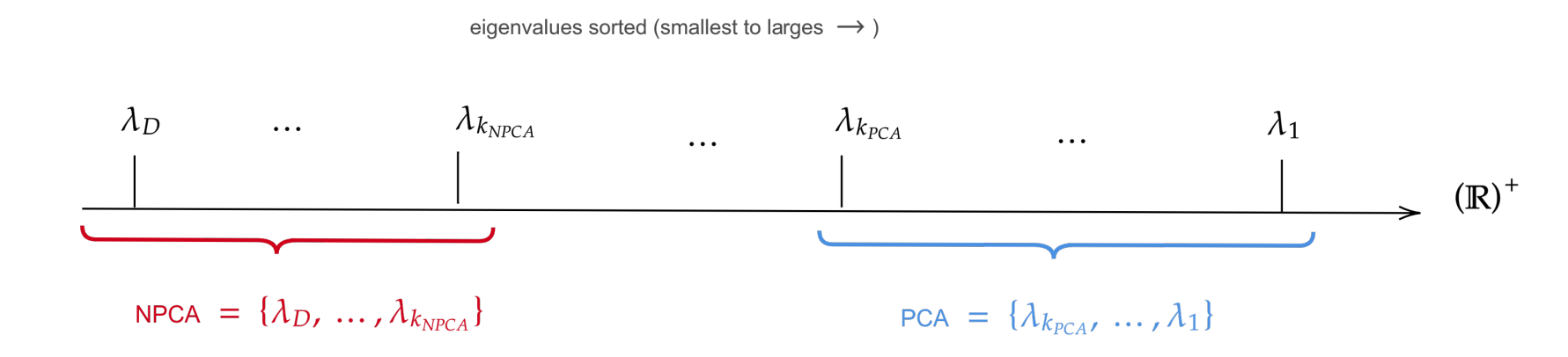}
\caption{PCA and NPCA algorithm}
\label{fig:1}
\end{figure}

While in AD task, the defects are always so subtle that we propose to do the Negated Principal Component Analysis (NPCA). We select the matrix consisting of the eigenvectors corresponding to the k features with the smallest eigenvalues (i.e., the smallest variance), represented in Fig.3, thus transforming the original matrix into the new space.

\subsection{Whitening transformation}
We propose another way to calculate the score of classifier anomolies by applying the whitening transformation on the features,
\begin{equation}
\mathbf{x}_{whitening} = \Sigma^{-\frac{1}{2}} (\mathbf{x}-\mu)
\label{eq}
\end{equation}
with $\Sigma^{-\frac{1}{2}}$ being square root of inverse of covariance matrix $\Sigma^{-1}$. 
And then calculate the euclidean distance of whitened features, which should be the same as the mahalanobis distance of original features.

After this whitening transformation, the distribution of normal features transforms to the standard normal distribution so we could do the normality test on each axis of features in the experiment below.

\section{experiment}

Usually, a large-scale neural network could have
stronger representation ability and extract more kinds of feature maps with its numerous layers. EfficientNet has been proved to achieve a
better performance than ResNet but with a much
shallower architecture and fewer parameters \cite{b6}. Following the same experimental settings as \cite{b3}, we employ EfficientNet-B4 as the
feature extractors. And Table.1 shows the number of channels on each stage of layer for the architecture. 

Then we propose two heuristics and do the experiments with them.
\begin{table}[htbp]
	\centering
	\caption{Number of Channels in EfficientNet-B4}
	\label{tab:1}  
	\begin{tabular}{ccccc ccccc}
		\hline\hline\noalign{\smallskip}	
		stage & 0 &1 & 2 & 3 & 4 & 5 & 6 &7 &8  \\
		\noalign{\smallskip}\hline\noalign{\smallskip}
		Number & 48 & 24 &32 &56 &112 &160 &272 &448 &1792 \\
		
		\noalign{\smallskip}\hline
	\end{tabular}
\end{table}

\subsection{Normality test of whitened operation}
In statistics, normality tests are used to determine if a data set is well-modeled by a normal distribution and to compute how likely it is for a random variable underlying the data set to be normally distributed. Since the transformation to a standard normal distribution after the whitening operation, higher the p-value of test, more likely it conforms to that distribution. So our hypothesis is that $\tilde{k}$ may be obtained at the maximum value of p-value.

We applied the Kolmogorov–Smirnov test (K-S test) because it works if the mean and the variance of the normal are assumed known under the null hypothesis. When we focus on the feature for one category extracted from one stage of layers $ l \in \{0, 1, 2, \dots , 8 \} $ and apply the whitening transformation, there is the matrix of whitened features,
\begin{equation}
F_l=
\begin{pmatrix}
f_{11} & f_{12}& \cdots & f_{1d} \\
f_{21} & f_{22}& \cdots & f_{2d} \\ 
\vdots & \vdots&\ddots & \vdots \\
f_{n1} & f_{n2}& \cdots & f_{nd}
\end{pmatrix}
\label{eq}
\end{equation}
where $d$ is the number of components of the whitened feature and n is the number of samples in one certain category.
Then we apply NPCA with $k \in \{1, 2, \dots, d \}$ on the features and compute the mean of p-value $P_k$,
\begin{equation}
P_k =\frac{1}{k} \sum_{i=1}^k p (f_{1i} , f_{2i} , \dots {f_{ni}} )
\label{eq}
\end{equation}
where $p$ is the function of p-value of K-S test on one axis $ (f_{1i} , f_{2i} , \dots {f_{ni}} )$ for all the samples in this category

\subsection{Relative distance from synthetic images}
In AD task, the training set contains only normal images because in real industrial production, we cannot predict the possible abnormal types and the number of such defective items is very small. Therefore, we artificially create some abnormal images based on normal ones using the same generation process than DRAEM \cite{b7}. One sample of huzelnut is showed in Fig.4.

\begin{figure}[!h]
\centering
\includegraphics[scale=0.6]{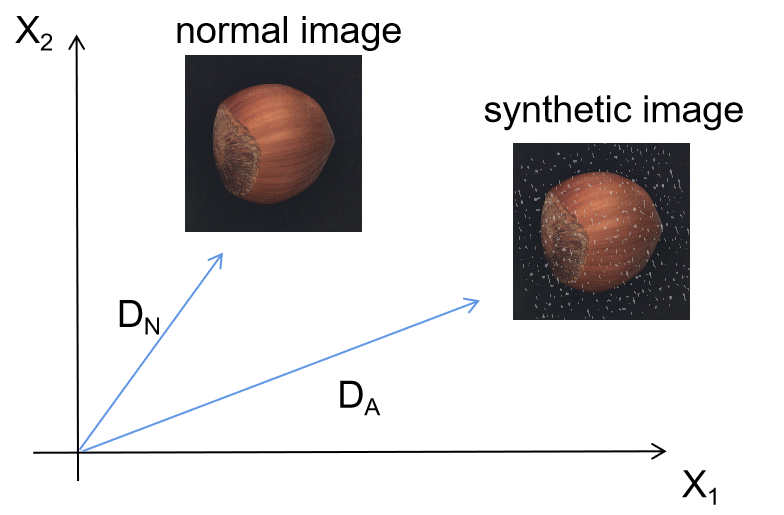}
\caption{A pair of augmentation images in hazelnut category}
\label{fig:1}
\end{figure}

Also, if we focus on the feature for one category extracted from one stage of layers $l$, the matrix of whitened features $F_l$ has the same dimension $n \times d$ with the matrix of whitened features of synthetic image $F_{Sl}$. 
We do the NPCA with all possible parameters $k \in \{1, 2, \dots, d \}$ and calculate the corresponding $R_k$.

Firstly we compute the score $s_{Ni} = \frac{1}{k} \sum_{i=1}^k f_{1i}^{2}$ of each normal image $i \in \{1, 2, \dots, n \}$. So we have the matrix of scores $S_N$ for normal images,
\begin{equation}
S_N =\begin{pmatrix}
s_{N1} & s_{N2} & \cdots & s_{Nn}
\end{pmatrix}
\label{eq}
\end{equation}
and we calculate the matrix of scores $S_A$ for synthetic images in the same way,
\begin{equation}
S_A = \begin{pmatrix}
s_{A1} & s_{A2} & \cdots & s_{An}
\end{pmatrix}
\label{eq}
\end{equation}

Secondly we calculate the relative increasing score for each pair of images and have the relative matrix $\frac{S_A}{S_N}$,
\begin{equation}
\frac{S_A}{S_N} = \begin{pmatrix}
\frac{s_{A1}}{s_{N1}}&\frac{s_{A2}}{s_{N2}}& \cdots & \frac{s_{An}}{s_{Nn}}
\end{pmatrix}
=\begin{pmatrix}
r_1 & r_2 & \cdots & r_n
\end{pmatrix}
\label{eq}
\end{equation}

Finally we can get the relative increasing score $R_k = \sum_{j=1}^n r_j$.

\section{result}
Contrasted to the mean auroc of all categories with certain parameters $k$ of NPCA in \cite{b3}, we generalize the auroc of all single category with all possible $k$ for each layer.
\begin{figure}[!h]
\centering
\includegraphics[scale=0.2]{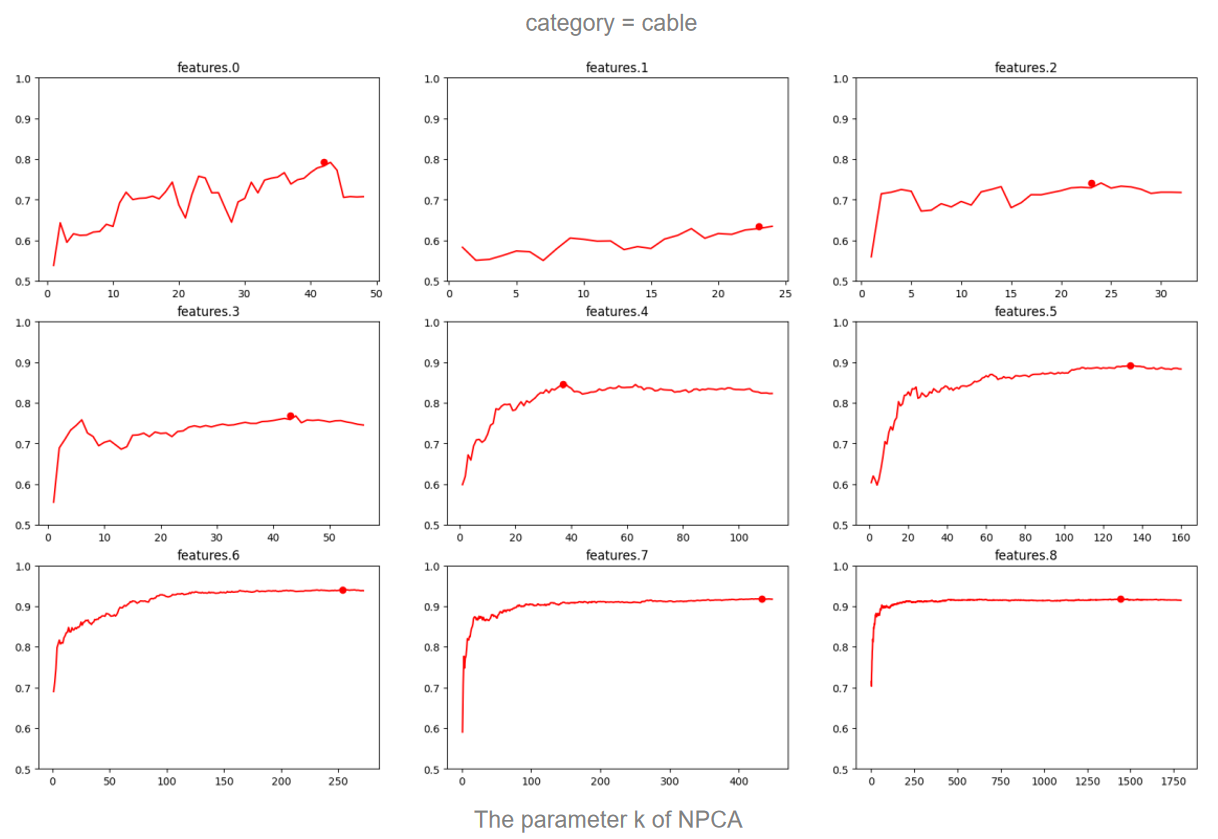}
\caption{auroc of cable across all layers}
\label{fig:1}
\end{figure}
In Fig.5, the red point represents the maximum auroc on each stage of layers for category cable. We could find that it exists an optimal $k^*$ when auroc achieves the maximum but not strictly locates at the last point (considering all components of feature). We wish the heuristics could help find $k^*$.

\subsection{Ratio of eigenvalues heuristic}
We reproduce the curve of heuristic in \cite{b3}. Fig.6 represents the example of feature.5 across all categories. The auroc curve fits well with the auroc in some categories, like tile and zipper. While the curve and auroc divergence in some categories, like carpet and grid, so the choice of $k$ in this heuristic tends to be bigger than the real $k^*$.
\begin{figure}[!h]
\centering
\includegraphics[scale=0.2]{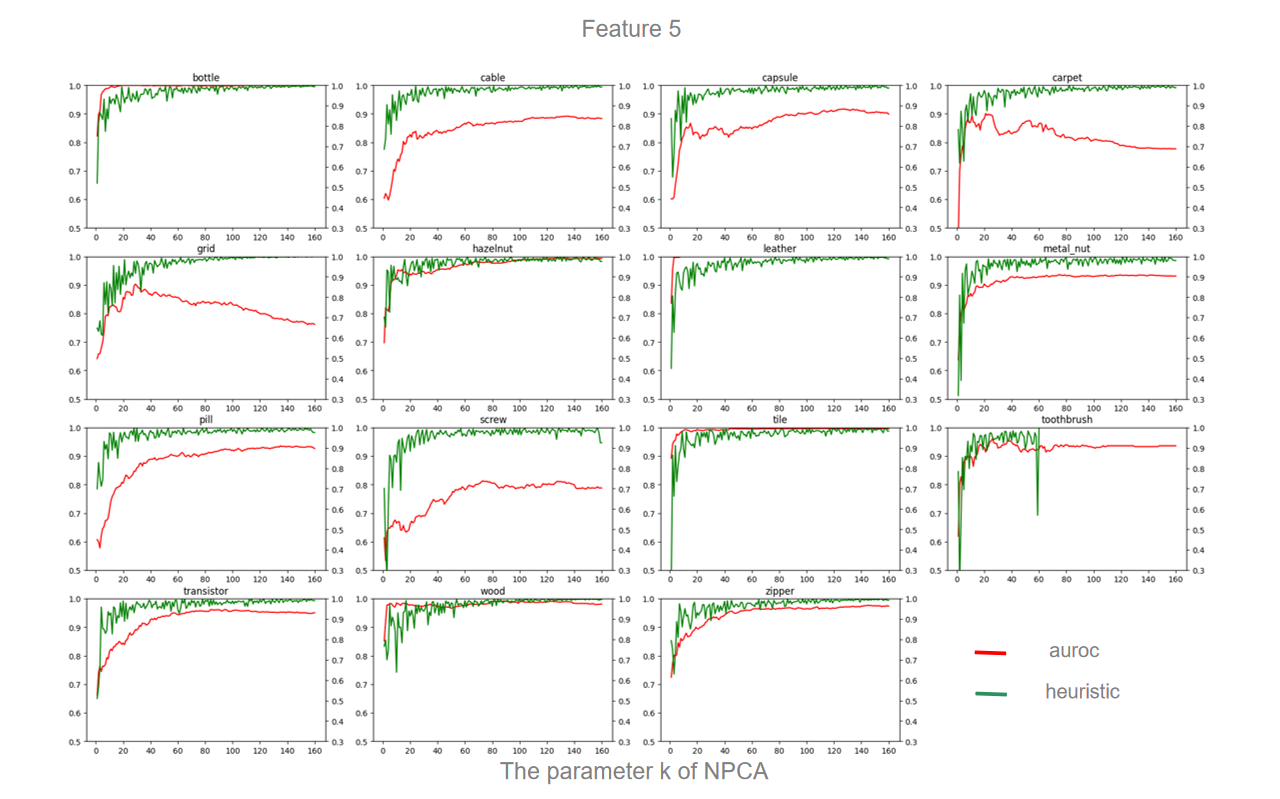}
\caption{Ratio of eigenvalues heuristic of feature5 across all categories}
\label{fig:1}
\end{figure}

\subsection{Normality test heuristic}
After doing the whitening transformation on the features extracted, we can do the normality test calculating $P$. In Fig.7, it shows the example of feature.6 across all categories.

\begin{figure}[!h]
\centering
\includegraphics[scale=0.2]{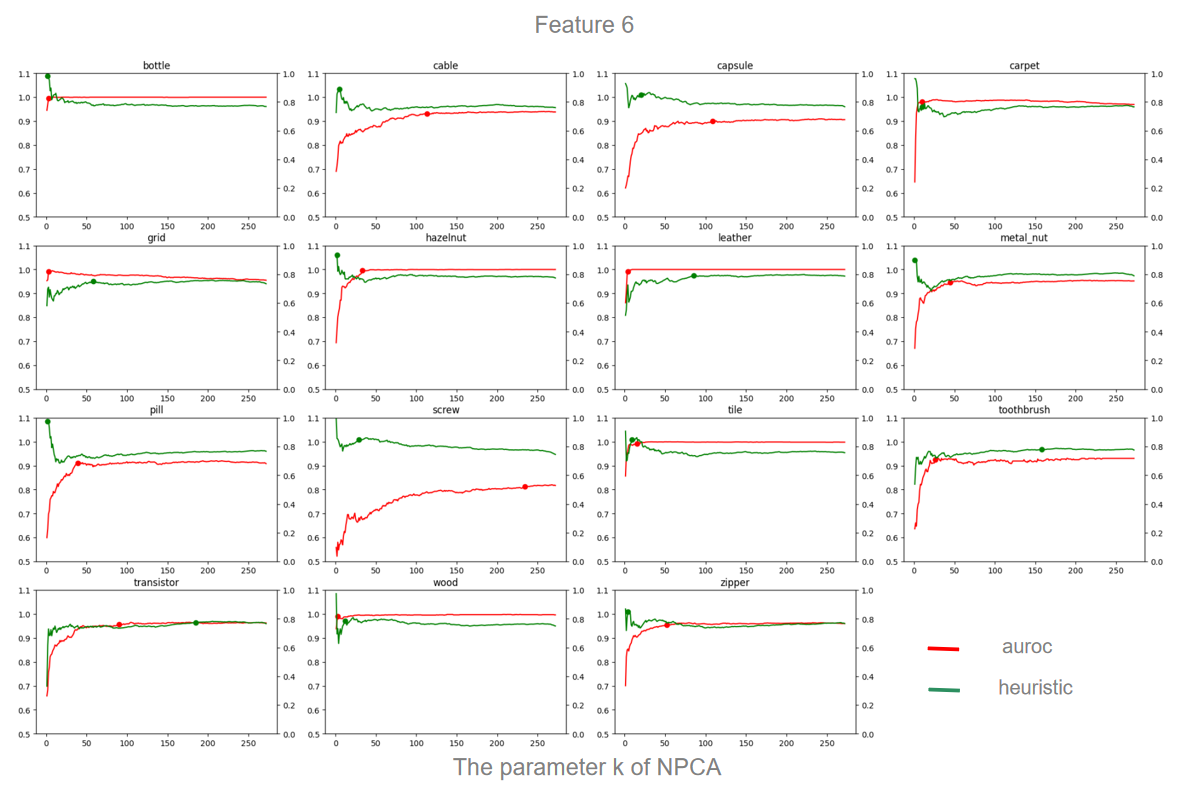}
\caption{Normality test heuristic of feature6 across all categories}
\label{fig:1}
\end{figure}

Here, we propose the method to choose $\tilde{k}$. If the maximum value of heuristic curve appears on the first point ($\tilde{k} = 1$), we discard it because it is generally not the real $k^*$. Then we try to determine the $\tilde{k}$ from the first minimum point of the curve. 

Because we want to pick the smallest possible value of $\tilde{k}$, so we set the tolerance of $1 \%$ of the maximum value on both curves and choose the minimal $\tilde{k}$ whose corresponding value enters this range. 

In Fig.7, the choice of $\tilde{k}$ (green point) conforms approximately to $k^*$ (red point) for carpet and tile categories. But the difference is large for other categories.

\begin{figure}[!h]
\centering
\includegraphics[scale=0.2]{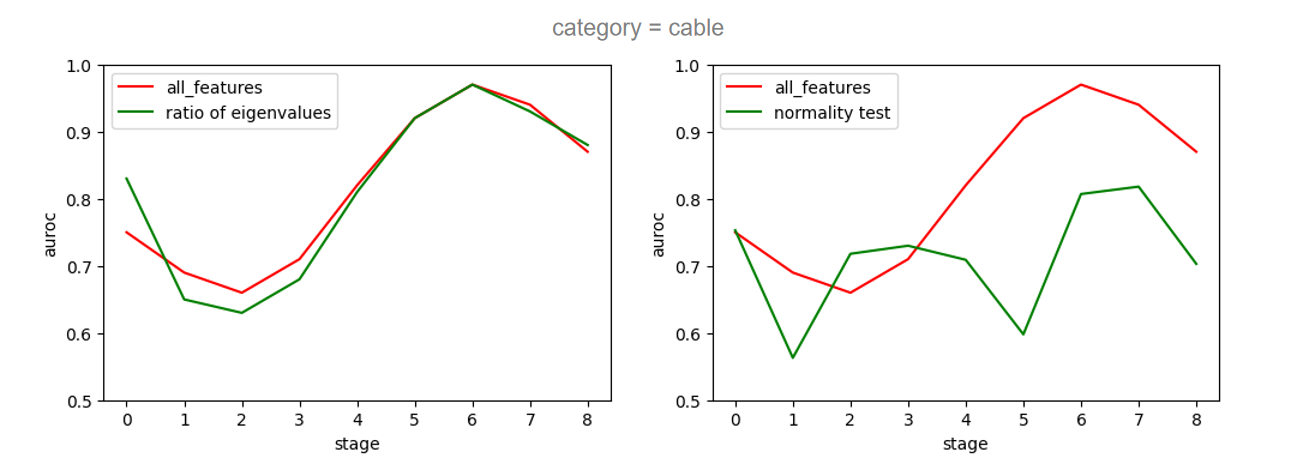}
\caption{Comparison of two heuristics}
\label{fig:1}
\end{figure}

We tried to compare this two heuristics showing the example of category cable across stage of layers, represented in Fig.89. After choosing parameter $\tilde{k}$ of NPCA for each feature and corresponding auroc, the ratio of eigenvalues heuristic performs much better, which fits well with the auroc of all features. The normality test heuristic actually loses much information for anomaly detection.

\subsection{Relative distance heuristic}
In Fig.9, in addition to the curve of relative distance heuristic represented at left, we then compute its differential curve at right because we have noted an obvious turning point at the heuristic curve. So we get a better fitting curve.

\begin{figure}[!h]
\centering
\includegraphics[scale=0.45]{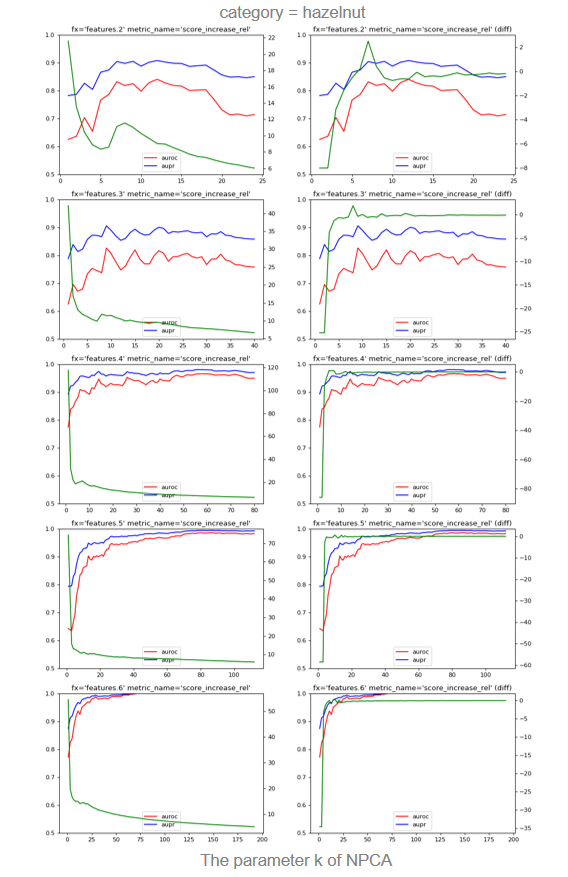}
\caption{Relative distance heuristic of hazelnut from feature2 to feature6}
\label{fig:1}
\end{figure}

For the example of hazelnut from feature.2 to feature.6 in Fig.9, the differential curve of relative distance with respect to parameter $k$ fits well with the auroc curve. So this augmentation of normal images seems to gives an estimate of a good choice of $k$ and offers the direction of our future work. 
\\ 

In addition to the above three heuristics, we also did experiments for four other heuristic, but the results were very bad. Accumulated eigenvalues and pseudorank \cite{b4} seem to be very rough signal to choose $k$, Muirhead's method \cite{b4} leads to a uncomputable curve because it tends to infinity and Levina-Bickel's algorithm \cite{b5} need tune the parameter for each case which is another issue to study.

\section{conclusion}
We have proposed some heuristic to choose the parameter of NPCA, but none of them fits perfectly with the auroc. In comparison, still the ratio of eigenvalues heuristic provides a more stable selection of features.
But on the other hand, we obtained some simple synthetic images by augmentation of the normal images. Studying the relative distance of these images and corresponding real normal images, we found it performs better than our heuristic that takes account only the normal ones. So this is a more promising heuristic of hyperparameter choice for anomaly detection.

\end{document}